\documentclass[conference]{IEEEtran}
\IEEEoverridecommandlockouts

\pdfoutput=1

\ifCLASSINFOpdf
   \usepackage[pdftex]{graphicx}
\else
\fi
\usepackage{amsmath}
\usepackage{ amssymb }

\usepackage{caption}
\usepackage{subcaption}

\usepackage{url}

\usepackage{color}
\usepackage{multirow}
\usepackage{rotating}

\begin{document}
\bstctlcite{IEEEexample:BSTcontrol}

\title{A Context Aware and Video-Based Risk Descriptor for Cyclists}

\author{Miguel Costa$^{1}$ and Beatriz Quintino Ferreira$^{2}$ and Manuel Marques$^{3}$
\thanks{$^{1}$Miguel Costa is a Master student with the ECE department,
        Instituto Superior Tecnico, 1049 Lisboa, Portugal
        {\tt\small miguel.n.costa@tecnico.ulisboa.pt}}%
\thanks{$^{2}$ Beatriz Quintino Ferreira is a Ph.D student with the ECE department,
        Instituto Superior Tecnico, 1049 Lisboa, Portugal
        {\tt\small beatriz.quintino@tecnico.ulisboa.pt}}%
\thanks{$^{3}$ Manuel Marques is a Researcher with the ECE department,  Instituto Superior Tecnico, 1049 Lisboa, Portugal
        {\tt\small manuel@isr.ist.utl.pt}}%
}


\maketitle

\begin{abstract}
Aiming to reduce pollutant emissions, bicycles are regaining popularity specially in urban areas. However, the number of cyclists' fatalities is not showing the same decreasing trend as the other traffic groups.
Hence, monitoring cyclists' data appears as a keystone to foster urban cyclists' safety by helping urban planners to design safer cyclist routes.
In this work, we propose a fully image-based framework to assess the route risk from the cyclist perspective. From smartphone sequences of images, this generic framework is able to automatically identify events considering different risk criteria based on the cyclist's motion and object detection. Moreover, since it is entirely based on images, our method provides context on the situation and is independent from the expertise level of the cyclist. 
Additionally, we build on an existing platform and introduce several improvements on its mobile app to acquire smartphone sensor data, including video. 
From the inertial sensor data, we automatically detect the route segments performed by bicycle, applying behavior analysis techniques.  
We test our methods on real data, attaining very promising results in terms of risk classification, according to two different criteria, and behavior analysis accuracy.     
\end{abstract}


%
\IEEEpeerreviewmaketitle

\section{Introduction}
Bicycles are winning back importance in our society as a sustainable means of transportation, specially in urban areas~\cite{EU-reporttraffic:2016,Strauss:deceleration:2017}. In fact, not only do they bring positive impact to the environment, but also to public health and traffic~\cite{Gossling:TransportTransitions:2013}. 
Both the European Union and the USA are committed to raise the number of cyclists while increasing cycling safety~\cite{Pucher:CyclingIrres:2008}. Notwithstanding, we have witnessed a much lower decrease (3\%) in the number of cyclist fatalities when compared to the fatalities reduction in the other traffic groups (around 18\%)~\cite{EU-roadsafety:2015}.  

Collecting traffic data, in particular cyclists' data, is very important for urban planners and a keystone to design safer cycling routes. 

With the advent of smartphones and other mobile wearable devices, acquiring massive sensory data for behavior analysis has become not only highly affordable but also a common practice~\cite{Vlahogianni:DrivingSurvey:2017}; so these appear as a perfect match to this task.

In this work we explore this synergy: use sensor data to assess the route risk, fostering safety and mobility for urban cyclists.

In fact, several recent studies have been focusing on collecting and analyzing different types of traffic data (video, GPS, acceleration, orientation) in an attempt to evaluate and improve road operation and safety regarding cyclists~\cite{Strauss:deceleration:2017,Cara:Car-CyclistClass:2015,Muralidharan:HR-trafficsensing:2014}, as well as promote active commuting, which leads to reduction of air pollution and congestion in traffic networks~\cite{Hasshu:ActiveCommuting:2015}.   

\begin{figure}[t]
  \centering
  \includegraphics[width=0.65\columnwidth]{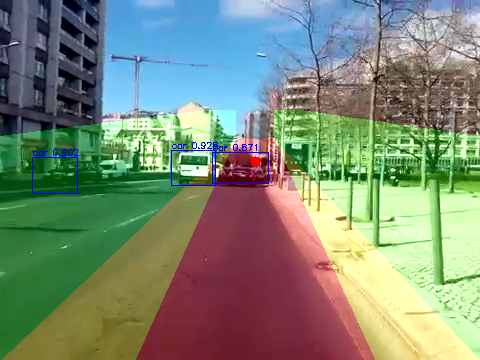}
  \caption{Risk analysis descriptor: The estimation of the Focus of Expansion (red point) enables to define three risk zones (red, yellow and green); our risk descriptor is correlated with the occupancy of each zone by detected objects (blue rectangles). In this image, the detection of a car in the red zone (cyclist' route) indicates a possible high risk situation. Differently, the car in the green zone represents a lower risk for the cyclist.}
  \label{fig:intr}
  \vspace{-5mm}
\end{figure} 

In particular, we believe that monitoring cyclists route risk can be valuable to improve safety as well as help guide city planning. 
Our previous work, the SMARTcycling tool presented in~\cite{Vieira:SMARTcycling:2016}, was the first to automatically identify generic driving events that may condition cyclists' real commuting experience. In the latter work, stressful events were detected using a bio-metric sensor.
Although the method was able to assign stress levels to segments of the paths performed by the cyclists, it required a posteriori visual inspection of the acquired images in order to understand what particular event generated the stress level variation (e.g. other passing by cyclists, cars or pedestrians, road anomalies, to name a few).  
Moreover, it was observed that similar patterns were obtained during stress and effort situations (e.g. due to terrain elevation), requiring disambiguation through image and GPS data analysis. We may also postulate that the identification of stressful events based on biological signals may be user-dependent, varying accordingly to the cyclist's experience, comfort, or physiological characteristics, among other factors.

We center our approach on smartphone data, taking advantage of the video captured by our mobile application (dubbed Bike Monitor) to develop an alternative method based on optical flow and focus of expansion (FOE) to assess risky events from the external factors from the route, providing also the context for each situation (see Figure~\ref{fig:intr}). 

Therefore, we build on the SMARTcycling tool from~\cite{Vieira:SMARTcycling:2016}, proposing a novel fully image-based method to assess the cyclist's route risk, which is also context and motion aware. 
Additionally, we introduce significant improvements in the Bike Monitor app towards a more exhaustive and reliable data acquisition, including performing behavior analysis so that we only analyze the segments of the route in which the user is actually riding a bicycle (as opposed to walking or riding a motorized vehicle). 
Furthermore, relying solely on the cyclist's smartphone image and sensor data is a step towards a cyclist invariant method.

To the best of our knowledge, this is the first automatic method to identify, contextualize (using images), and assess dangerous riding events for cyclists entirely based on smartphone data.
More information about the SMARTcycling project and the Bike Monitor app can be found at: \url{http://users.isr.ist.utl.pt/~manuel/smartbike/}.

We highlight the following contributions of our work:
\begin{itemize}
\item Image-based and context-aware assessment framework of dangerous events (based on semantic and optical flow descriptors), described in Section \ref{sec:ImageProc};
\item Behavior analysis based on smartphone sensor data, automatically delimiting the portions of the path performed riding a bicycle
, as described in Section \ref{sec:BehavAnal}. 
\end{itemize}
Moreover, and preceding the previous contributions, we introduce improvements on the SMARTcycling tool. Specifically, we develop new features for the Bike Monitor app, mainly at the back-end layer, but also: user profile registration, video acquisition/upload from the smartphone camera, report and registration of performed routes in a map allowing post inspection. This is described in Section \ref{sec:app};

\section{Related Work}
In the past few years, sensing human activity has become ubiquitous and traffic has been no exception. In this vein, several studies have focused on collecting traffic data to monitor road conditions~\cite{Gonzalez:RDS:2017,Seraj:RoadAnomalyDetec:2015}, roadway operation~\cite{Muralidharan:HR-trafficsensing:2014,Panichpapiboon:TafficSensing:2016}, and assessing driving experience~\cite{Johnson:DrivingStyle:2011,Araujo:DrivingCoach:2012,Eren:DrivingBehavior:2012}. However, the large majority of these works~\cite{Gonzalez:RDS:2017,Araujo:DrivingCoach:2012,Johnson:DrivingStyle:2011,Eren:DrivingBehavior:2012,Seraj:RoadAnomalyDetec:2015} target motorized vehicles as these are still dominant in today's traffic volume. 
Nevertheless, bicycle usage has recently grown mainly in urban areas~\cite{Strauss:deceleration:2017}; and perhaps the marginal decrease of cyclists' fatality in comparison to all other road groups~\cite{EU-roadsafety:2015} is a by-product of this trend. These facts have raised awareness to cyclists' safety, and the research community is starting to give more and more attention to this issue (see, for example, the February 2017 Safety Science special issue on Cycling Safety).  

Compared to motorized vehicles, collecting and processing cyclist data is more challenging, as bicycles are less stable (no suspension) resulting in noisier data. This lack of stability is specially problematic when processing images acquired by a smartphone attached either to the bicycle or the cyclist.   

Cara et al.~\cite{Cara:Car-CyclistClass:2015} circumvent this issue by using an instrumented car to acquire data in order to classify car-cyclist scenarios. In this work the authors test machine learning algorithms on bicycle-car interaction data to classify safety-critical scenarios, envisaging the development of Advanced Driver Assistance Systems (ADAS) that support cyclist protection.

Despite the aforementioned technical hurdles, some recent studies address cycling experience by equipping bicycles with on-board sensors. Aiming at specialized cycling intelligent systems,~\cite{Dozza:DynamicsBehavior:2014} proposes a framework to understand bicycle dynamics and cyclist behavior. Such framework and collected data can be seen as an important pre-requisite to the development of bicycle suited applications.     

Concerning cyclists' safety,~\cite{Strauss:Mappingactivityrisk:2015} and~\cite{Jacobsen:safetyinnumbers:2003} study the relation between the number of cyclists going through a given lane or intersection and the risk of crash with other cyclist and motorist, respectively.
More recently, Strauss et al.~\cite{Strauss:deceleration:2017} use a large sample of GPS cyclists' trip data acquired via a smartphone application in order to validate deceleration rate as a surrogate safety measure. Particularly, the authors explore the correlation of deceleration with accidents at intersections as a potential proactive measure to prevent cyclist injuries.

Yet, in terms of the sensors used, there has been practically no distinction between assessing drivers' or cyclists' experience, as previous methods usually depend on inertial sensor data (such as accelerometer or gyroscope).

In~\cite{Vieira:SMARTcycling:2016} we introduced a new approach to detect and identify driving events primarily based on processing images from an action camera. We are able to overcome the issues of using a camera mounted on the bicycle as an acquisition sensor, since the natural shake of the cyclist’s movement is filtered at the computation of the optical flow.
In its previous version, the SMARTcycling tool captured and processed data from the cyclist's smartphone, an action camera, and a cardio acquisition belt. Applying image processing techniques based on optical flow descriptors to the action camera videos, the SMARTcycling tool showed good accuracy on driving events classification and road condition identification. This tool was also able to evaluate cyclists' stress using the ECG data collected from a bio-metric belt.   

Due to its amenable properties, in terms of set-up and data acquisition, we claimed that SMARTcycling~\cite{Vieira:SMARTcycling:2016} paves the way to large scale assessment, as cities often provide public bicycle sharing programs, where it can be easily deployed.

In this work we delve into more involved computer vision and image processing techniques to be able to automatically identify and contextualize dangerous events from external factors, sparing both the action camera and the bio-metric belt, which imply a more complex set-up. 
The descriptor used in~\cite{Vieira:SMARTcycling:2016} was context independent (splitting the image into fixed zones), we now use a different and richer approach that encodes the context surrounding the cyclist when performing event detection. 
Moreover, contrary to our previous work, we analyze the whole image, incorporating motion, temporal dependence and image semantics. 

Regarding semantics, Aly et al.~\cite{Aly:MapCrowdSensing:2016} propose an approach to crowd-sense users' smartphones to automatically enrich digital maps with semantic road information such as road condition, bridges or crosswalks. However, and once again, the proposed algorithms only rely on inertial sensor measurements. 

Here we follow a different direction, taking advantage of the good properties yielded by using images as primary source of data.
Indeed, computer vision techniques have been applied, for quite some time, to traditional cyclist monitoring tasks as volume counts~\cite{heikkila:realtime_monitoring:2004} and average speed, due to their reliability and efficiency when compared to manual methods~\cite{Zaki:CV:CollectionCyclistData:2013}. 
However, so far no work has addressed identification of dangerous situations using an on-board smartphone camera.

We apply state-of-the-art classification methods (convolutional Neural Networks, specifically the Faster R-CNN from~\cite{ren:faster-RCNN:2015}) to obtain the localization and presence probability of objects in the image. The semantics provided by object detection and classification allows to interpret and understand the detected dangerous situations, providing much more insight than other types of measurements. 



\section{Bike Monitor App}
\label{sec:app}
As introduced in~\cite{Vieira:SMARTcycling:2016}, the SMARTcycling tool has its own smartphone data acquisition interface - the BikeMonitor app.

Bike Monitor runs on Android operative system and has a very simple and intuitive interface that allows user profile registration, start and stop data recording, and upload the recorded data to the server. 

Upon registration, the user is asked to provide her/his age, gender, cycling experience level and bicycle characteristics (suspension/no suspension). This data is stored in the server and since it is organized by user account the profile is automatically associated with new uploads from the same user.

In this new version, we further explore the rich sensing capabilities of today's smartphones, adding the recording of the following signals: speed (from GPS), linear and gravitational acceleration along the three axes (X,Y,Z), rotation matrix, orientation, and GPS uncertainty.
The interval between acquisitions is now 0.1s (was 0.5s), and all signals are indexed by a time-stamp, allowing a time synchronized processing.

In addition to inertial sensors and GPS data, the Bike Monitor app has now the option to record video and sound from the smartphone camera and microphone, respectively.  
Bearing in mind battery life issues, video acquisition is configurable in terms of quality (low or high) and frequency (1, 5 or 30 fps). 

After uploading the recordings from each journey, the user receives an automatically generated map summarizing the ride.


\section{Image-based risk assessment}
\label{sec:ImageProc}

In order to provide a framework to detect and assess risky events for cyclists, we use video sequences from the smartphone camera and combine different computer vision techniques to obtain descriptors based on optical flow and semantics. In particular, we start by estimating the FOE to embed the cyclist motion into the descriptor. We then compute a risk descriptor that considers the objects present in the image and the division into zones according to the estimated FOE. Finally, we can assess risk considering different criteria, by using the obtained descriptor and computing a specific distance metric for the specified criteria.  

\paragraph{Estimating the FOE}
Differently from~\cite{Vieira:SMARTcycling:2016} and before computing the optical flow, we split the image into 16 zones and filter each zone with the histogram equalization method (CLAHE)~\cite{Pizer:CLAHE:1987} to enhance contrast and edges definition.
We apply the Shi-Tomasi corner detector~\cite{Shi-Tom:CornerDetect:1994} and the feature extraction method from~\cite{Grundmann:RollingShutter:2012} to find sufficient and evenly distributed points of interest, even in regions with low texture. 

\begin{figure}
  \centering
  \includegraphics[width=0.65\columnwidth]{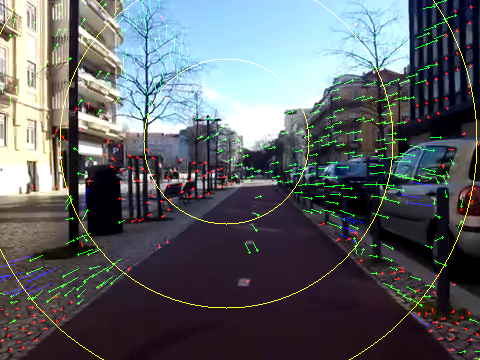}
  \caption{Optical Flow vectors and associated weights: red corresponds to $m_i = 0.1$, blue $m_i = 0.75$ and green $m_i = 1$.}
  \label{fig:OF_magnitude}
  \vspace{-5mm}
\end{figure} 

The optical flow vector $v_i$ on the image point $p_i$ is computed applying the Lucas and Kanade algorithm~\cite{Lucas:OpticalFlow:1981}. With the optical flow vectors, our goal is to compute the focus of expansion (FOE): a single point in the scene where all the velocity vectors meet. 
To improve robustness to outliers, we incorporate spatio-temporal prior knowledge about the optical flow and FOE in our 2D images. As Figure~\ref{fig:OF_magnitude} shows, the magnitude of the optical flow vectors increases with the distance to the FOE~\cite{Lee:est_OFV:2009} and we exploit this fact to iteratively perform outlier rejection.
According to Figure~\ref{fig:OF_magnitude}, we first divide the image into 4 concentric circles centered around the previous calculated FOE. We then calculate the distribution of the optical flow vectors magnitude in each annulus (formed by the circle excluding its inner circles) and on the innermost circle. Given the average magnitude of its zone $\overline{v_{\Omega_i}}$, each optical flow vector $v_i$ has an associated magnitude weight $m_i$, according to the following expression:
\newcommand*\rfrac[2]{{}^{#1}\!/_{#2}}
\vspace{-2mm}
\begin{equation}
\label{eq:OF_weight_magnitude}
\begin{aligned}
m_i = \begin{cases}
 0.10 \text{ , if } & \text{abs}(\|v_i\|-\overline{v_{\Omega_i}}) \geq (\overline{v_{\Omega_i}})^\frac{2}{3} \\ 
 0.75 \text{ , if } & (\overline{v_{\Omega_i}})^\frac{1}{2} < \text{abs}(\|v_i\|-\overline{v_{\Omega_i}}) < (\overline{v_{\Omega_i}})^\frac{2}{3} \\ 
 1.00 \text{ , if } & \text{abs}(\|v_i\|-\overline{v_{\Omega_i}}) \leq (\overline{v_{\Omega_i}})^\frac{1}{2}\\ 
\end{cases}
\text{,}
\end{aligned}
\end{equation}
where $\overline{v_{\Omega_i}}=\frac{\sum_{j\in\Omega_i}\|v_j\|}{|\Omega_i|}$, $\Omega_i$ is the set of indices whose optical flow vectors $v_j$ are in the same annulus of $v_i$ and $\text{abs}(a)=|a|$.


The previous process is only valid for static scenes. However, this is not the case for traffic images, as though their background is static there are objects moving.
In order to discover the non-static points, we feed the Faster R-CNN~\cite{ren:faster-RCNN:2015} with each image frame to find the class and location (given as a bounding box) of the objects present. Since the probability that the detected objects are static is low (our classes of interest are persons, bicycles and ground motorized vehicles), we weight the flow vectors associated with each object by the negative exponential of the confidence score $s$ output by the neural network. Equation \eqref{eq:OF_weight_objects} shows the weight $o_i$ assigned for each optical flow vector $v_i$\footnote{Note that when an object is detected at point $p_i$, the score $s_i$ for that point coincides with the score $s_l$ for the detected object.}. 
\begin{equation}
\label{eq:OF_weight_objects}
\begin{aligned}
o_i = e^{-s_i}.
\end{aligned}
\end{equation}
This way we minimize the impact of the flow vectors associated with points with high probability of being objects. On the other hand, if $v_i$ is not associated with any object its weight is maximum $o_i=1$ because $s_i=0$. Hence, we take advantage of the image semantics to reduce the FOE estimate error. 

Considering these two types of weights, each optical flow vector $v_i$ has an associated weight given by
\begin{equation}
\label{eq:OF_vector_final_weight}
\begin{aligned}
w_i = m_i \cdot o_i.
\end{aligned}
\end{equation}

Computing $w_i$ for each optical flow vector $v_i$, we can estimate the FOE in a non-static scenario. In light of the FOE definition, this is equivalent to finding the closest point to a set of $N$ lines (extensions of the optical flow vectors). Although this problem can be solved via Least-Squares, we estimate the solution point using the Huber Loss~\cite{Hastie:ElementsSL}, as it deemphasizes outliers.   
Let us define $f(x,L_i)$ as the distance between a point $x \in \mathbf{R}^2$ and a line $L_i$, parameterized by $L_i = \{p_i + tu_i : t \in \mathbf{R}\}, p_i, u_i \in \mathbf{R}^2, u_i = \frac{v_i}{\|v_i\|} $ for $i=1,...,N$, as $f(x,L_i) = \sqrt{(x-p_i)^T(I-u_iu_i^T)(x-p_i)}$.   
We formulate and solve the following optimization problem 
\begin{equation}
\label{eq:huber-loss_opt_prob}
\begin{aligned}
& \tilde{x} = \underset{x}{\text{argmin}}
& & \sum_{i=1}^N \mathcal{L}_{\delta}\left(\frac{1}{w_i} \cdot f(x,L_i) \right), 
\end{aligned}
\end{equation}  
where $\mathcal{L}_{\delta}(a)$ is the Huber Loss given by
\begin{equation}
\label{eq:huber-loss_}
\begin{aligned}
& \mathcal{L}_{\delta}(a) =
& & \begin{cases} 
      \frac{1}{2} a^2, & |a|\leq \delta \\
      \delta(|a|-\frac{1}{2}\delta), & otherwise. \\
   \end{cases}
\end{aligned}
\end{equation}   
Solving \eqref{eq:huber-loss_opt_prob} (with $\delta = 1$) we find the point that minimizes the sum of weighted distances $\frac{1}{w_i} \cdot f(x,L_i)$ for all the obtained optical flow vectors, penalized by the Huber Loss. 

Similarly to the previous static case, we perform an iterative refinement of the weighted lines $L_i$ that are considered in this computation. Specifically, we solve problem \eqref{eq:huber-loss_opt_prob} and then remove the optical flow vectors whose orientation is not according to the optimal FOE found. We repeat this process, solving \eqref{eq:huber-loss_opt_prob} for a new weight assignment, until it converges, i. e., the difference between the FOE estimates in two consecutive iterations is smaller than a predefined threshold or a maximum number of iterations is achieved.

Exploring the smoothness of the cyclist's trajectory, we perform a weighted average with the FOE of the current and $M$ previous frames as
\vspace{-2mm}
\begin{equation}
\label{eq:compute_FOE_avg}
\begin{aligned}
x_t = \frac{\sum\limits_{j=t-M}^{t} \tilde{x}_j \cdot e^{-\tau(t-j)}}{\sum\limits_{j=t-M}^{t}e^{-\tau(t-j)}},
\end{aligned}
\end{equation}  
where $x_t$ is the FOE estimate at instant $t$, $\tilde{x}$ is the minimizer of \eqref{eq:huber-loss_opt_prob} at the time instant $j$ and $\tau$ the decay rate of the weights.

\begin{figure}
  \centering
  \includegraphics[width=0.65\columnwidth]{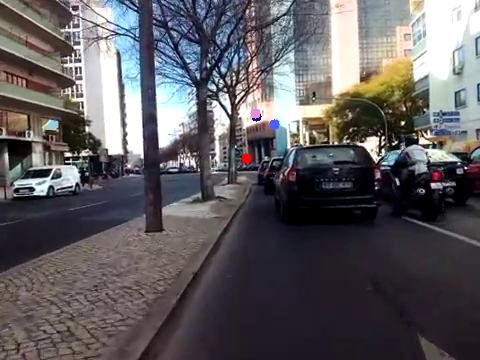}
    \vspace{-2mm}
  \caption{Intermediate and final (in red) FOE estimates. Light blue shows the estimate given by the Huber Loss without weights, dark blue the Huber Loss estimate considering weights $w_i$, and pink the estimate after the iterative refinement.}
  \label{fig:FOE_points}
  \vspace{-5mm}
\end{figure}

Figure \ref{fig:FOE_points} illustrates the intermediate estimates (until convergence) and final FOE (shown in red).

\begin{figure}
  \centering
  \includegraphics[width=0.65\columnwidth]{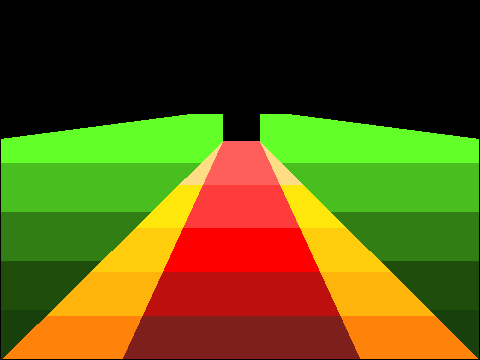}
    \vspace{-2mm}
  \caption{Division of the 5 different risk zones into 25 sub-regions to promote proximity encoding.}
  \label{fig:zonesOcc}
  \vspace{-5mm}
\end{figure}

\paragraph{Computing the risk descriptor}
The obtained FOE gives an estimate of the direction of the cyclist's movement. Based on this direction we can divide the image into five main regions according to the proximity to the cyclist's trajectory. Figures \ref{fig:intr} and \ref{fig:zonesOcc} show these regions, with a color code (red representing the region including the cyclist's predicted trajectory, yellow the region closest to the trajectory, and green the region farther away from this trajectory). 
The only assumption we make when dividing the image into these regions is that the camera is placed not too far from the ground level, approximately perpendicular to the motion direction, and is not facing up (to the sky).

In order to have a descriptor more spatially fine-grained, we subdivide horizontally each of the previous regions in 5 sub-regions, yielding a total of 25 sub-regions (see Figure \ref{fig:zonesOcc}) as the following expression of the descriptor at instant $t$ shows
\begin{equation}
\label{eq:descriptor}
d_t = \begin{bmatrix} d_t^1 & d_t^2 & \cdots & d_t^{24} & d_t^{25} \end{bmatrix}. 
\end{equation}  
The division into these sub-regions encodes the proximity to the cyclist depending on his motion. 

Given these motion and proximity aware sub-regions as well as scene object classification and location, we provide a framework to assess dangerous events that can use descriptors based on several criteria: lane occupation, proximity, type of passing by vehicles or combinations of these.  

We compute the risk score of each sub-region $k$ at instant $t$ as:
\begin{equation}
\label{eq:risk_subregion}
  \vspace{-2mm}
d_t^k = \sum_{l=1}^{N_t^k} r_t^{k,l} 
\end{equation}
where $N_t^k$ is the number of objects in sub-region $k$ at instant $t$. The risk associated to object $l$ in sub-region $k$ and instant $t$ is given by
\begin{equation}
\label{eq:risk_score_object}
r_t^{k,l} = \alpha_l\cdot s_l\cdot\gamma_k\cdot\frac{a_t^{k,l}}{b_t^k}
\end{equation}
where $\alpha_l$ is the object coefficient depending of its type (person, bicycle, car, etc.), $s_l$ the confidence score output by the neural network, $\gamma_k$ the coefficient of region and sub-region $k$, $a_t^{k,l}$ the area of object $l$ in sub-region $k$ and $b_t^k$ the area of sub-region $k$, both at instant $t$. Note that $\gamma_k$ codifies the 25 sub-regions and takes into account the larger five regions depicted in Figure~\ref{fig:zonesOcc}.
Expression \eqref{eq:risk_score_object} combines the fact that different objects (weighted by the classification confidence score output by the neural network) pose different risk levels, and that risk depends on both the cyclist's trajectory and object proximity (given by the regions and sub-regions). Also, the ratio between the area occupied by each object and the total sub-region area informs on how close and how large each object is. 

\paragraph{Computing distance metric for the descriptor}
At this point, our risk assessment framework outputs a risk score for each image sub-region. To provide a more informative assessment and easier to understand by the user, we propose to encode these risk scores in a single global risk level.  

We formulate this as a supervised classification problem, and use the Earth Mover's Distance (EMD) metric to perform image retrieval and classify new images in each class~\cite{Rubner:EMD:2000}. EMD is known to match well perceptual similarities for image retrieval when compared to other distances~\cite{Rubner:EMD:2000}. 
If we have intrinsic relations between distribution bins, EMD is a measure of the distance between two distributions and finding the minimum cost that has to be paid to transform one distribution into the other can be cast as a transportation problem. Such formulation is a linear optimization problem for which efficient algorithms are available~\cite{Rubner:EMD:2000}. 

In our case, to compare risk events, we wish that sub-regions that are close in the image and belong to the same risk level have small distance, that the distance between two sub-regions increases with the image distance between them and with risk level dissimilarity. Also, we wish to have a small distance between sub-regions that are symmetric in the image. Then, we design a $25 \times 25$ distance matrix which assigns distance values between all pairs of sub-regions.  

Different descriptors (based on different criteria) can be specified by defining a scale of global risk levels and designing a ground distance matrix (which is an input of the EMD image retrieval) that better models the relation of the criteria and the image locations (sub-regions).  
In our experimental results (see Section \ref{sec:results}) we instantiate this framework, assessing risk based on two separate criteria: lane occupation and proximity.


\section{Behavior analysis}
\label{sec:BehavAnal}
Given our intent of creating mobility profiles for the users, we seek to automatically classify the type of transportation taken in each part of a route, sparing user input. For that, we use a supervised learning approach, which relies on labeled data provided by the Bike Monitor app (collected from real users, under real-world circumstances without researcher supervision).
Specifically, we use Support Vector Machines (SVMs) as they are a widely used method, very flexible, fast and efficient, do not have many parameters to tune~\cite{Hastie:ElementsSL}.

In this section we describe all the steps of our human activity classification \textquotedblleft pipeline", from data acquisition and preprocessing to feature selection. 
Specifically, after collecting the dataset we preprocess the signals (cleaning and windowing) before extracting relevant features. Once the features are extracted we can perform classification. To maximize accuracy, we also add temporal continuity to the classification~\cite{Krishnan2008}, due to the continuous nature of the activities in study. 

In order to keep a low computational cost without compromising accuracy, we adopt the following strategies: extract the majority of features from time domain signals, choose SVM classifier (known to be computationally efficient), and implement a feature selection method~\cite{Guyon2002} that can drastically reduce the number of features used by the SVM.
 
We detail the steps of our classification approach below.

\paragraph{Data acquisition and preprocessing}
We use the Bike Monitor App to collect the signals from the smartphone's sensors (see Section \ref{sec:app} and~\cite{Vieira:SMARTcycling:2016}).
We selected the following signals: linear acceleration along the three axes (X, Y, Z), gyroscope data to compute rotations also along (X, Y, Z), and GPS data to obtain speed.  
We discard the first and last 10 seconds of each signal to avoid mislabeling, as during these periods the user may be still setting up for the activity or may be already stopped~\cite{Bao2004}.

When selecting a time window it is fundamental that it is long enough to contain the whole activity under analysis, and, on the other hand, short enough that it does not include additional events.  
Previous works on activity recognition report good accuracy results with sliding windows covering approximately 5 to 10 seconds of movement. Considering  our scenario, the app sampling frequency and implementation constraints, features are computed on sliding windows of 100 samples (with a 50\% overlap, as this overlap percentage has been successful in the past~\cite{Bao2004}).

\paragraph{Feature Extraction}
Feature extraction is a critical step in the design of any classifier. 
We explore the following statistical features previously used in the literature~\cite{Bao2004,Krishnan2008}: mean, standard deviation, root-mean square and mean absolute deviation. In addition to the latter time domain analysis, we extract some frequency domain features using the Fast Fourier Transform, computing the power spectral entropy and spectral energy for each window.
   
Furthermore,~\cite{Bao2004} reports that features measuring correlation of acceleration between axes can improve recognition of activities involving  multiple body parts. Thus, we also include features encoding the correlation between all pairs of axes. 

Finally, we obtain, for each window, a feature vector with a total of 54 features (including time and frequency domain features computed from the 3-axes acceleration, gyroscope and GPS speed signals and the time domain features of the acceleration cross correlation between pairs of axes).
Grouping all feature vectors results in the predictor data matrix $\mathbf{X}$. 

\paragraph{Classification Method - SVM}
We use SVMs to classify human activity, based on the previous features, into three classes: cycling, walking and riding a motorized transport (e.g. a car or a bus).


As maximal margin classifiers, SVMs are widely used, benefit from computational advantages over probabilistic methods and are known to perform well on high dimensional data~\cite{Murphy:2012:MLP}. 
Although originally designed for binary classification, the One-Versus-All (OVA) and One-Versus-One (OVO) are possible approaches to extend SVMs to multi-class problems~\cite{Murphy:2012:MLP}.  

Kernels allow to extend SVMs to cases where the datasets are not linearly separable. This is achieved by kernel functions which translate the original data to a new space, using basis expansions such as polynomials or splines~\cite{Hastie:ElementsSL}. 

\paragraph{Adding Temporal Continuity and Feature Selection}
Although SVMs are effective in classifying individual frames, they do not account for temporal continuity~\cite{Krishnan2008}. 
To this end, we add the generic framework proposed in~\cite{Krishnan2008} for incorporating temporal continuity for classification of continuous human activity on top of our SVM classifier.
The underlying idea is that probability values computed for a frame at time instant $i$ ($f_i$) can benefit the classification of successive temporally close frames.  
Specifically, the probability of a frame $f_t$ belonging to class $c$ is weighted on the temporal distance and similarity between current and past frames. This induces more recent frames to have more impact in the current frame than older ones and assumes that if adjacent frames are identical, then they should belong to the same class  (see~\cite{Krishnan2008} for details). We add this temporal continuity to the whole set of signals in our dataset.  


To keep the classification cost low and prevent overfitting it is important to select relevant features. 
In this vein, we apply the technique introduced in~\cite{Guyon2002} for feature pruning specifically for SVM, based on Recursive Feature Elimination (RFE). In a nutshell, RFE iteratively trains the classifier and computes a ranking criterion for all features, removing the feature with lowest ranking. Applying RFE to our SVM classifier we are able to significantly reduce the dimensionality of our problem (as we will see in Section \ref{sec:results}).

\section{Experimental Results}
\label{sec:results}

\subsection{Image-based risk assessment}
We tested our risk classification approach for a total of approximately 300 labeled image frames (with close to 100 frames belonging to each of the three risk levels), acquired by the Bike Monitor app by real users.
We split our image dataset into training and test set, according to a 75\% to 25\% ratio.
 
\begin{figure}
\centering
\begin{subfigure}{.24\textwidth}
  \centering
  \includegraphics[width=\textwidth]{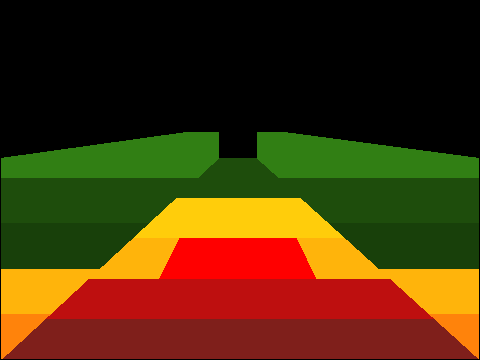}
  \vspace{-5mm}
  \label{fig:zonesProxExemple}
\end{subfigure}
\begin{subfigure}{.24\textwidth}
  \centering
  \includegraphics[width=\textwidth]{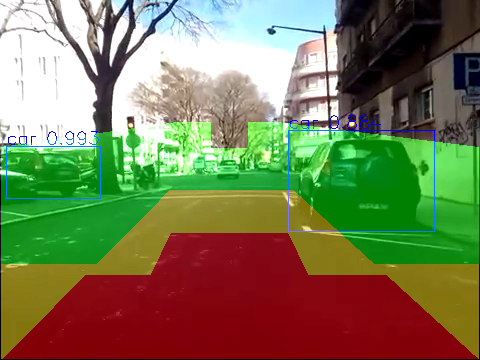}
  \vspace{-5mm}
  \label{fig:zonesProx}
\end{subfigure}
\vspace{-1mm}
\caption{Proximity risk classifier. a) Proximity regions encoding: red corresponds to the highest risk, yellow to intermediate and green to the lowest risk; b) Proximity based risk regions superimposed on a RGB image.}
\label{fig:proximityCriteria}
\vspace{-5mm}
\end{figure}

As we claimed earlier, our risk assessment framework is general and can be applied to different criteria. Here we show results for classifiers based on lane occupation and proximity. In the former we study the risk associated with the path occupation or trajectory of the user and define the regions as in Figures \ref{fig:intr} and \ref{fig:zonesOcc}, whereas in the latter we assess the risk associated with the proximity of objects to the cyclist, and define the risk regions as shown in Figure \ref{fig:proximityCriteria}.

We manually labeled each image according to the levels defined for the two different criteria.
In order to assess cyclists' risk based on different criteria one must define risk levels according to the specific criterion and design a distance matrix (used by the EMD) that properly captures the intended notion of nearness. 

We define three risk levels (higher level means higher risk). For both classifiers, we consider the risk levels as: 3- the red region is occupied; 2- the yellow region is occupied, and; 1- only the green region is occupied.
The distance matrices used for the two classifiers follow the desired properties outlined above. Notwithstanding, we add some alterations to better fit each criterion. 
The distance matrix used together with the lane occupation criterion adds a multiplicative factor ($> 1$) to distances between sub-regions belonging to different risk regions (represented by red, yellow or green). On the other hand, for the proximity criterion we add a multiplicative factor (also $> 1$) when the sub-regions belong to different semi-circular zones, centered on the cyclist.

To estimate the optical flow vectors we used the Lucas and Kanade algorithm~\cite{Lucas:OpticalFlow:1981} with squared windows of 35 pixels, 1 pyramid level, and a 5 skip frame. These parameters depend on the resolution and frame rate; in our experiments we used a resolution of 480$\times$360 and 30fps.

To compute the risk score for each object (see equation \eqref{eq:risk_score_object}) we consider that motorized objects (cars, buses and motorcycles) present higher risk than bicycles which in turn present higher risk than persons. Hence, we assign a higher object type value (1) for motorized vehicles, an intermediate value (0.8) for bicycles, and a lower value (0.6) for person detections. 
The region risk is defined according to the region to each sub-region belongs to: being higher for sub-regions belonging to the red region, intermediate for sub-regions in the yellow region and lower for sub-regions within the green region.
The sub-region risk decreases as we move up vertically in the image (sub-regions near the bottom present higher risk than sub-regions near the top). 
As object area, instead of directly using the bounding box provided by the Faster-RCNN detection, we computed the area ratio considering the width given by the bounding box, and the height (in pixels) as $max\{\mbox{0.2} \times \mbox{bounding box height}, 10\}$. This area is a better approximation of the object projection in the defined risk zones (which map regions on the ground), as we consider that all discoverable objects classes are in contact with the ground. All previous variables belong to the interval $[0,1]$, yielding a $\mbox{risk}_{\mbox{score}}(j)$ for each object also between 0 and 1.  

Moreover, we use the Python Toolbox \emph{sklearn}~\cite{scikit-learn} to solve the optimization problem in \eqref{eq:huber-loss_opt_prob} to estimate the FOE, and the EMD implementation from the \emph{pyemd}~\cite{pele:pyEMD:2009}.

\begin{table}[!tb]
   \caption{Risk Classification}
    \begin{subtable}{.45\linewidth}
          \centering
      \caption{Lane occupation based}
      \label{table:laneOccupation_confusion_matrix}

        \begin{tabular}{cc|l|l|l|}
\cline{3-5}
\multicolumn{2}{c}{\multirow{2}{*}{}}                     												& \multicolumn{3}{|c|}{Predicted Class} \\ \cline{3-5} 
\multicolumn{2}{c}{}                                       				   & \multicolumn{1}{|c|}{1} & \multicolumn{1}{c|}{2} & \multicolumn{1}{c|}{3} \\ \hline
\multicolumn{1}{|c|}{\multirow{3}{*}{\rotatebox[origin=c]{90}{\begin{minipage}{0.25in}Class\end{minipage}}}} 											& \multicolumn{1}{c|}{1} & \multicolumn{1}{c|}{80} & \multicolumn{1}{c|}{20}  & \multicolumn{1}{c|}{0}  \\ \cline{2-5} 
\multicolumn{1}{|c|}{}                                 															& \multicolumn{1}{c|}{2} & \multicolumn{1}{c|}{9.1} & \multicolumn{1}{c|}{81.8}& \multicolumn{1}{c|}{9.1} \\ \cline{2-5} 
\multicolumn{1}{|c|}{}                                 															& \multicolumn{1}{c|}{3} & \multicolumn{1}{c|}{0} & \multicolumn{1}{c|}{25}  & \multicolumn{1}{c|}{75} \\ \hline
\end{tabular}
    \end{subtable}%
    \begin{subtable}{.45\linewidth}
      \centering
        \caption{Proximity based}
        \label{table:proximity_confusion_matrix}
        \begin{tabular}{cc|l|l|l|}
\cline{3-5}
\multicolumn{2}{c}{\multirow{2}{*}{}}                     												& \multicolumn{3}{|c|}{Predicted Class} \\ \cline{3-5} 
\multicolumn{2}{c}{}                                       				   & \multicolumn{1}{|c|}{1} & \multicolumn{1}{c|}{2} & \multicolumn{1}{c|}{3} \\ \hline
\multicolumn{1}{|c|}{\multirow{3}{*}{\rotatebox[origin=c]{90} {\begin{minipage}{0.25in}Class\end{minipage}}}} 											& \multicolumn{1}{c|}{1} & \multicolumn{1}{c|}{66.7} & \multicolumn{1}{c|}{33.3}  & \multicolumn{1}{c|}{0}  \\ \cline{2-5} 
\multicolumn{1}{|c|}{}                                 															& \multicolumn{1}{c|}{2} & \multicolumn{1}{c|}{10.7} & \multicolumn{1}{c|}{82.1}& \multicolumn{1}{c|}{7.2} \\ \cline{2-5} 
\multicolumn{1}{|c|}{}                                 															& \multicolumn{1}{c|}{3} & \multicolumn{1}{c|}{0} & \multicolumn{1}{c|}{41.2}  & \multicolumn{1}{c|}{58.8} \\ \hline
\end{tabular}
    \end{subtable} 
      \vspace{-3mm}
\end{table}

We show the results of our risk classification as a confusion matrix in Table \ref{table:laneOccupation_confusion_matrix} for the Lane Occupation Risk classifier, and in Table \ref{table:proximity_confusion_matrix} for the Proximity Risk classifier. 
We note that there is no misclassification between risk levels 1 and 3 in any of the classifiers. Thus both classifiers separate well these two extreme classes. 
The achieved accuracy for the Lane Occupation classifier is relatively high, showing an error rate of 20-25\% for each class. For the Proximity classifier, results show some missclassification between risk levels 3 and 2, which we deem to be a result of objects that appear close to the limits of both red and yellow zones upon labeling, and thus incurring some error in the classification. Furthermore, as our risk levels are not continuous, i.e., we have discretized the risk levels throughout the defined areas and not used a smooth continuous risk function, it is expected that the risk classification incurs in some errors when objects are positioned close to the boundaries of each zone.


\subsection{Behavior Analysis}
We divided our dataset for behavior analysis (with a duration of approximately 8 hours) keeping again a ratio of approximately 75\% of training to 25\% of test data~\cite{Hastie:ElementsSL}. 

Accuracy is evaluated based on a loss function measuring the classification error for the SVM model, computed using the test examples and the corresponding true class labels. 
The used loss function is given by
$L = \sum_{i=1}^N a_i I\{\hat{y_i} \neq y_i\}$, 
where $a_i$ is the weight of observation $i$ (these weights sum to the respective class prior probability, which are normalized so that all priors sum to one), $I(x)$ is the indicator function, $\hat{y_i}$ is the class label given by the SVM as the class with the maximal posterior probability, and $y_i$ the true class label. 

Table \ref{tab:Loss_SVM_OVA} shows the average loss obtained for different kernel functions and parameters $C$ (penalization imposed to points violating the SVM margin).  

\begin{table}[]
\centering
\caption{Classification error loss for OVA SVM-classification}
\label{tab:Loss_SVM_OVA}
\begin{tabular}{lc|c|c|c|c|}
\cline{3-6}
                                               &                & \multicolumn{4}{c|}{$C$} \\ \cline{3-6} 
                                               &                & 0.5     & 1  & 10 & 20 \\ \hline
\multicolumn{1}{|c|}{\multirow{4}{*}{\rotatebox[origin=c]{90}{Kernels}}} & Linear         & 0.0118       & 0.0091  & 0.0754  & 0.0783  \\ \cline{2-6} 
\multicolumn{1}{|c|}{}                         & Gaussian       & 0.6011  & 0.5951   & 0.5951   & 0.5951   \\ \cline{2-6} 
\multicolumn{1}{|c|}{}                         & Polyn. Order 2 &    0.0236     &  0.0236  &  0.0236  &  0.0236  \\ \cline{2-6} 
\multicolumn{1}{|c|}{}                         & Polyn. Order 3 &   0.0266      & 0.0266   & 0.0266   &  0.0266  \\  \hline
\end{tabular}
  \vspace{-0mm}
\end{table}

The SVM classifier achieves highest accuracy (approximately $99\%$) for $C=1$ and a linear kernel.

To maximize accuracy, we incorporate temporal continuity by feeding the score of the SVM as input to the method of~\cite{Krishnan2008}.  
We added temporal continuity to the cases that attained higher classification accuracy for the previous \textquotedblleft temporal insensitive" SVMs. 
Table \ref{tab:add_temp_OVA} presents the results obtained. 

\begin{table}[]
\centering
\caption{Classification error loss when adding temporal continuity}
\label{tab:add_temp_OVA}
\begin{tabular}{c|c|c|}
\cline{2-3}
                                                   & No temporal cont. & Adding temporal cont. \\ \hline
\multicolumn{1}{|c|}{Linear} & 0.0091                 & 0.0091                     \\ \hline
\multicolumn{1}{|c|}{Polyn. Order 2} & 0.0236                 & 0.0168                     \\ \hline
\end{tabular}
  \vspace{-5mm}
\end{table}

Adding temporal continuity maintains the highest attained classification accuracy (for the linear kernel), but it increases (by an order of 30\%) the accuracy for the Polynomial kernel.


The previous accuracy results were obtained considering the full set of 54 features. Yet, we can reduce the problem dimensionality by applying the feature selection technique SVM-RFE from~\cite{Guyon2002}.
Although this method was proposed for the binary case only, we start by selecting features in the $K=3$ binary classifiers and then we experiment the multi-class case with the most relevant (highest ranked) features found before.
With this list of ranked features one can study the impact on the achieved accuracy of eliminating less relevant features, as well as understand what are the most discriminative features (by trying to grasp some intuitive physical interpretation).   

Training and testing the previous SVM that maximized accuracy for OVA (with $C=1$ and linear kernel) and inputting only the 8 most relevant features found (by performing a kind of consensus between the binary classifiers) returns a loss of $0.0647$. Hence, we move from $\approx 99\%$ accuracy when including all 54 features to $\approx 94\%$ after reducing the dimension of the data to 8. This result is very promising, since we can significantly lower the computational load at the cost of a slight accuracy reduction.

We observed that when classifying between walking and another class the losses were very low, suggesting the walking class has very distinctive features with respect to the others. 
Contrastingly, a significant accuracy degradation was observed when pruning features for the Bike\emph{Vs}Car classifier.

\begin{figure}
  \centering
  \includegraphics[width=0.74\columnwidth]{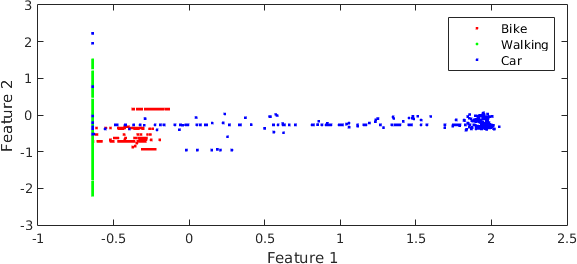}
    \vspace{-2mm}
  \caption{Scatter diagram of the data reduced to 2 dimensions}
  \label{fig:SVM_3}
  \vspace{-6mm}
\end{figure}

Reducing even more the data dimensionality allows us to visually grasp the multi-class problem. Figure \ref{fig:SVM_3} shows all training data reducing the predictor $\mathbf{X}$ to the root-mean square of the speed and the mean rotation along Y (the two most relevant features found by agreement of the 3 binary classifiers).  
We observe that speed (feature 1) is very effective to separate the classes (it shows remarkably low inner-class variance for Walking).
However, we also note that some Bike and Car observations are mixed (these two can originate similar speeds specially within the context of traffic jams).
 \vspace{-4mm}
\section{Conclusions}

In this work we contribute with a novel and complete framework to assess the risk of cyclists' routes. 
First, we performed improvements on an existing platform (a mobile app that recorded multiple smartphone sensor data), so that we could base our method entirely on videos acquired from the cyclists' smartphones. 
Then, we proposed a generic framework for image risk descriptors, based on optical flow (to compute the FOE) and semantics, thus being context-aware and invariant to the user. 
Additionally, we performed behavioral analysis based on smartphone sensor data to automatically detect when the user is riding a bicycle, as opposed to riding a motorized vehicle or walking.  
Combining several methods from computer vision, image processing, and statistical learning, we were able to overcome many technical hurdles that are common when acquiring and dealing with cyclist's data.  
Instantiating this framework for two specific criteria (lane occupation and proximity), our risk assessment was shown to perform well for both cases on real data.
Similarly, the behavior analysis was also tested on real data and achieved very good accuracy results.  
Finally, considering this work's potential for city planning and road accidents prevention, we hope it can reach and influence our cities' decision makers.

We identify several possible directions for future work, including taking advantage of the good properties of deep neural networks in our classification problem and making our FOE estimates even more robust. 
Particularly, we envisage to train our own deep neural network only with images taken from a cyclist's point of view. This way we expect detection and classification of objects to have even higher accuracy. Also, we could benefit from deep learning to classify risk given the FOE and detected objects as input, bypassing the formulation as an image retrieval problem and using EMD.   
To improve robustness of the proposed FOE estimation method, we could use the dominant direction of each route to help us better prune the optical flow vectors used (as they are the single source of error).   

\vspace{-3.5mm}
\bibliographystyle{IEEEtran}

\bibliography{IEEEabrv,smartbike}

\end{document}